\documentclass[preprint,12pt]{elsarticle}

\usepackage{amssymb}
\usepackage{amsmath}

\journal{Nuclear Physics B}

\begin{document}

\begin{frontmatter}



\title{Title: Understanding Image2Video Domain Shift in Food Segmentation: An Instance-level Analysis on Apples
}


\author[label1]{Keonvin Park\fnref{fn1}}
\author[label2]{Aditya Pal\fnref{fn1}}
\author[label3]{Jin Hong Mok\corref{cor1}}

\fntext[fn1]{These authors contributed equally to this work.}

\cortext[cor1]{Corresponding author. Email: jhmok1024@dgu.edu}

\affiliation[label1]{organization={Interdisciplinary Program in Artificial Intelligence, Seoul National University},
            addressline={1 Gwanak-ro, Gwanak-gu},
            city={Seoul},
            postcode={08826},
            country={Republic of Korea}}

\affiliation[label2]{organization={Department of Biological Environmental Science, College of Life Science and Biotechnology, Dongguk University},
            city={Seoul},
            postcode={04620},
            country={Republic of Korea}}

\affiliation[label3]{organization={ Department of Food Science and Biotechnology, Dongguk University},
            city={Goyang-si, Gyeonggi-do},
            postcode={10326},
            country={Republic of Korea}}

\ead{kbpark16@snu.ac.kr}
\ead{aditya@dgu.ac.kr}
\ead{jhmok1024@dgu.edu}

\begin{abstract}
Food segmentation models trained on static images have achieved strong performance on benchmark datasets; however, their reliability in video settings remains poorly understood. In real-world applications such as food monitoring and instance counting, segmentation outputs must be temporally consistent, yet image-trained models often break down when deployed on videos.
In this work, we analyze this failure through an \emph{instance segmentation and tracking} perspective, focusing on \emph{apples} as a representative food category. Models are trained solely on image-level food segmentation data and evaluated on video sequences using an instance segmentation with tracking-by-matching framework, enabling object-level temporal analysis.
Our results reveal that high frame-wise segmentation accuracy does not translate to stable instance identities over time. Temporal appearance variations, particularly illumination changes, specular reflections, and texture ambiguity, lead to mask flickering and identity fragmentation, resulting in significant errors in apple counting. These failures are largely overlooked by conventional image-based metrics, which substantially overestimate real-world video performance.
Beyond diagnosing the problem, we examine practical remedies that do not require full video supervision, including post-hoc temporal regularization and self-supervised temporal consistency objectives. Our findings suggest that the root cause of failure lies in image-centric training objectives that ignore temporal coherence, rather than model capacity. This study highlights a critical evaluation gap in food segmentation research and motivates temporally-aware learning and evaluation protocols for video-based food analysis.
\end{abstract}

\begin{graphicalabstract}
\includegraphics[width=\textwidth]{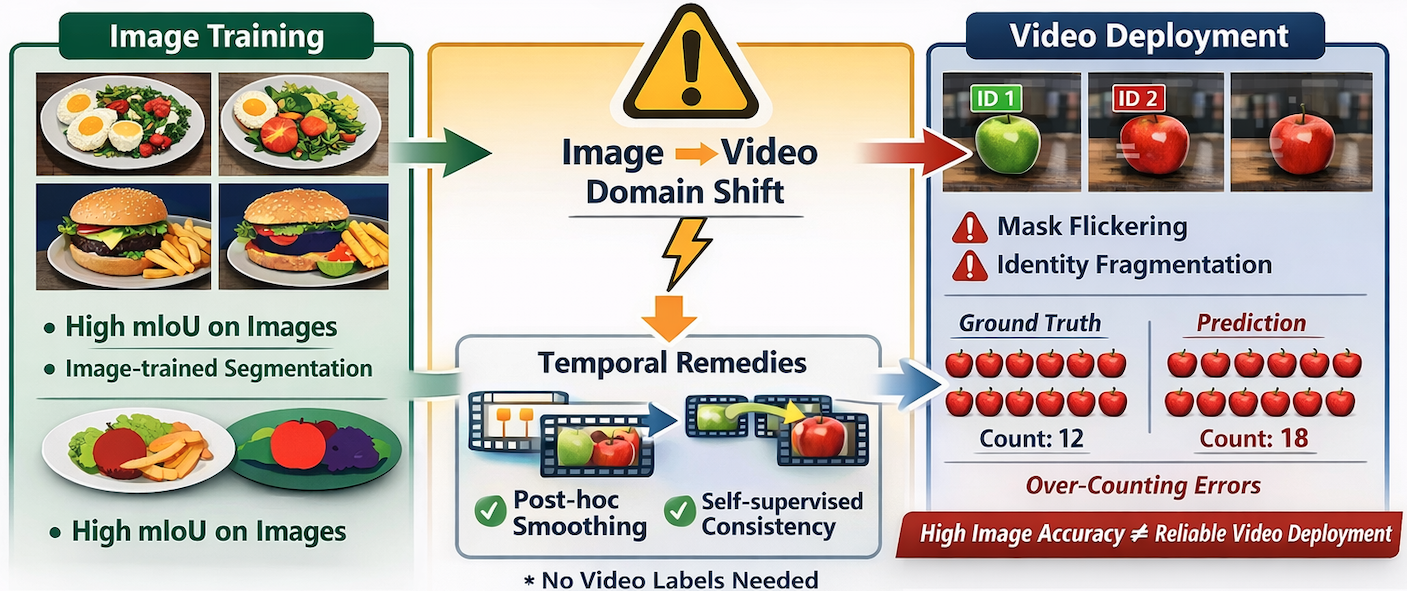}
\end{graphicalabstract}

\begin{highlights}
\item Analyzes why food segmentation models trained on static images fail in video scenarios.
\item Reformulates video food segmentation as an instance segmentation and tracking problem.
\item Reveals that high frame-wise accuracy does not ensure temporal consistency or reliable instance counting.
\item Identifies a critical evaluation gap between image-based benchmarks and video-based deployment.
\item Demonstrates that simple temporal consistency mechanisms can alleviate severe video failures.
\end{highlights}


\end{frontmatter}



\section{Introduction}
\label{sec1}

Food image analysis has become an essential component in applications such as dietary assessment, automated food monitoring, and intelligent food processing systems. Recent advances in deep learning have led to strong performance in food recognition and segmentation tasks, largely driven by large-scale image datasets and pretrained visual representations \cite{Bossard2014,FoodSurvey2025,Zhang2023}. As a result, food segmentation models trained on static images now achieve high accuracy on standard benchmarks.

However, most real-world food analysis systems operate on \emph{video streams} rather than isolated images. In these settings, segmentation outputs must remain consistent over time to support downstream tasks such as instance counting, portion estimation, and consumption tracking. Despite this requirement, food segmentation models trained exclusively on static images often exhibit severe performance degradation when applied to video data. Predictions that appear accurate on individual frames frequently suffer from mask flickering, identity fragmentation, and unstable object boundaries across frames.

This discrepancy reveals a fundamental \emph{image-to-video domain shift} that is largely overlooked in current food segmentation research. Conventional evaluation protocols rely on frame-wise metrics such as pixel-level mIoU, implicitly assuming temporal independence between frames. As a consequence, models with high image-level accuracy may still fail catastrophically when temporal consistency and instance identity are required. Similar limitations have been observed in broader studies on robustness and out-of-distribution generalization, where static benchmarks are shown to overestimate real-world performance \cite{Hendrycks2017,Hendrycks2019}.

From a representation learning perspective, image-trained models are optimized to capture spatial appearance cues while ignoring temporal coherence. In video, however, food appearance evolves continuously due to factors such as illumination variation, specular reflection, motion blur, and partial occlusion. These temporal appearance changes destabilize instance representations, leading to identity switches and fragmented tracks. While domain adaptation and generalization techniques have been extensively studied in image-based recognition \cite{Pan2010,Wang2018,DGSurvey2022}, their implications for video-based food segmentation remain underexplored.

In this work, we revisit food segmentation from a video-centric perspective by reformulating evaluation as an \emph{instance segmentation and tracking} problem. Rather than assessing predictions independently at each frame, we explicitly analyze whether segmented food instances can be consistently tracked over time. We focus on \emph{apples} as a representative food category due to their frequent occurrence, visual ambiguity, and relevance to counting-based applications. Models are trained solely on image-level food segmentation data and evaluated on video sequences using a tracking-by-matching framework, enabling object-level temporal analysis without requiring video annotations.

Our study reveals that high frame-wise segmentation accuracy does not guarantee stable instance identities. Temporal appearance inconsistency—particularly subtle changes in lighting and surface reflection—leads to mask flickering and identity fragmentation, resulting in significant errors in apple counting. Importantly, these failures are largely invisible to conventional image-based metrics, which substantially overestimate real-world video performance.

Beyond diagnosing the problem, we examine practical remedies that do not rely on full video supervision, including post-hoc temporal regularization and self-supervised temporal consistency objectives. Our findings suggest that the root cause of failure lies in \emph{image-centric training objectives} that ignore temporal coherence, rather than insufficient model capacity. By exposing this evaluation gap, our work highlights the necessity of temporally-aware learning and evaluation protocols for reliable video-based food segmentation.

\section{Related Work}
\label{sec:related}

\subsection{Food Segmentation}

Early research on food analysis primarily focused on image-level recognition and classification. With the introduction of pixel-wise annotated datasets such as UECFoodPix and FoodSeg103, food segmentation has received increasing attention as a fine-grained visual understanding task \cite{UECFoodPix,FoodSeg103}. Recent models achieve strong segmentation performance on static images, benefiting from advances in deep neural networks and large-scale pretraining. However, existing food segmentation benchmarks and evaluations remain image-centric and do not account for temporal consistency, which is critical in video-based applications.

\subsection{Video Instance Segmentation}

Video instance segmentation (VIS) aims to jointly segment and track object instances over time. Early approaches such as MaskTrack R-CNN associate per-frame instance masks using learned embeddings, while subsequent methods improve temporal association through cross-frame feature aggregation and memory mechanisms \cite{MaskTrackRCNN,CrossVIS,IFC}. Although VIS has been extensively studied in generic object domains, its application to food-related scenarios remains limited. In particular, the performance of image-trained food segmentation models under VIS formulations has not been systematically analyzed.

\subsection{Video Object Segmentation and Foundation Models}

Video object segmentation (VOS) methods such as STCN and XMem focus on class-agnostic mask propagation across frames \cite{STCN,XMem}. More recently, foundation models like Segment Anything have demonstrated strong generalization in image segmentation and have been extended to video settings \cite{SAM,SAM2}. Despite their impressive capabilities, these models often rely on strong spatial cues and may struggle with maintaining instance identity under appearance variation, especially in domain-specific settings such as food videos.

\subsection{Domain Shift and Robustness}

Distribution shift between training and deployment data is a well-known challenge in visual recognition. Prior work in transfer learning and domain generalization studies how representations learned from one domain fail to generalize to unseen distributions \cite{Pan2010,DGSurvey2022}. Robustness benchmarks further demonstrate that models achieving high benchmark accuracy can be fragile under realistic perturbations \cite{Hendrycks2019}. However, most robustness studies focus on static images, leaving the image-to-video domain shift in food segmentation largely unexplored.

\subsection{Evaluation Beyond Frame-wise Accuracy}

Standard evaluation protocols for segmentation rely on frame-wise metrics such as pixel-level mIoU, implicitly assuming temporal independence. In contrast, tracking and VIS literature emphasizes identity preservation and temporal consistency, using metrics originally developed for multi-object tracking \cite{MOTMetrics,VISMetrics}. This discrepancy highlights a fundamental evaluation gap when image-trained segmentation models are deployed in video settings. Our work bridges this gap by analyzing food segmentation failures through an instance segmentation and tracking perspective.

\section{Method}
\label{sec:method}

\subsection{Problem Setup}

We consider the problem of deploying food segmentation models trained on static images to video data.
Let $\mathcal{D}_{img}=\{(x_i, y_i)\}$ denote an image-level food segmentation dataset, where $x_i$ is a single image and $y_i$ is its corresponding pixel-wise annotation.
Models are trained solely on $\mathcal{D}_{img}$ using standard supervised segmentation objectives.

At inference time, the trained model is applied to a video sequence $\mathcal{V}=\{x_t\}_{t=1}^T$, where ground-truth video annotations are unavailable.
The objective is not only to produce accurate per-frame segmentation masks, but also to maintain consistent instance identities over time, which is essential for downstream tasks such as food instance counting.
This setting exposes the image-to-video domain shift inherent in food segmentation.

\subsection{Instance Segmentation and Tracking-by-Matching}

To evaluate temporal consistency, we reformulate video food segmentation as an \emph{instance segmentation and tracking} problem.
For each frame $x_t$, the image-trained segmentation model produces a set of instance masks
$\mathcal{M}_t = \{m_t^k\}_{k=1}^{N_t}$.

Temporal association is performed using a tracking-by-matching strategy commonly adopted in video instance segmentation \cite{MaskTrackRCNN,CrossVIS}.
Specifically, instance masks between consecutive frames are matched based on a combination of spatial overlap and appearance similarity.
For two instance masks $m_t^i$ and $m_{t+1}^j$, we compute an association score:
\begin{equation}
S(m_t^i, m_{t+1}^j) = \mathrm{IoU}(m_t^i, m_{t+1}^j),
\end{equation}
and assign identities using greedy matching with a fixed IoU threshold.
This simple association mechanism allows us to isolate temporal instability originating from the segmentation model itself, without introducing complex tracking modules.

\subsection{Temporal Failure Modes}

Under this formulation, we observe several recurrent temporal failure modes when applying image-trained food segmentation models to video data:
(i) \emph{mask flickering}, where instance masks fluctuate significantly across adjacent frames;
(ii) \emph{identity fragmentation}, where a single physical food item is assigned multiple identities over time; and
(iii) \emph{identity switching}, where identities are incorrectly exchanged between instances.
These failures often occur despite high frame-wise segmentation accuracy and are exacerbated by temporal appearance variations such as illumination changes, specular reflections, and partial occlusions.

\subsection{Evaluation Metrics}

Standard food segmentation evaluations rely on frame-wise metrics such as pixel-level mean Intersection-over-Union (mIoU).
While useful for assessing spatial accuracy, these metrics ignore temporal consistency.
To capture instance-level stability, we additionally report tracking-based metrics adapted from the multi-object tracking literature \cite{MOTMetrics}.
In particular, we measure identity switches and track fragmentation to quantify temporal instability.
This combined evaluation reveals discrepancies between frame-wise accuracy and temporal reliability.

\subsection{Practical Remedies without Video Supervision}

Beyond analysis, we examine lightweight remedies that do not require video-level annotations.
First, we apply \emph{post-hoc temporal regularization} by smoothing instance predictions across time using short-term temporal windows.
Second, we explore \emph{self-supervised temporal consistency} objectives, where predictions on adjacent frames are encouraged to be consistent without explicit labels.
These approaches are inspired by prior work on robustness and generalization under distribution shift \cite{Hendrycks2019,DGSurvey2022}.
Importantly, our goal is not to propose a new video segmentation architecture, but to demonstrate that temporal-aware mechanisms can partially mitigate failures rooted in image-centric training objectives.

\section{Data}
\label{sec:data}

\subsection{Training Data: Food Image Segmentation}

For training, we utilize publicly available image-level food segmentation datasets that contain pixel-wise annotations of food items. Specifically:

\begin{itemize}
    \item \textbf{FoodSeg103} \cite{FoodSeg103}: A large-scale dataset comprising over 14,000 food images with pixel-wise annotations across 103 classes. It contains diverse food types and background conditions captured under varying real-world scenarios.
    \item \textbf{UECFoodPix} \cite{UECFoodPix}: A dataset with fine-grained pixel-level annotations covering multiple food categories. It provides object-level masks suitable for instance segmentation tasks.
\end{itemize}

These datasets provide a diverse set of static images with detailed segmentation annotations, allowing models to learn spatial appearance cues of food items. While these datasets do not include video sequences, they form the basis of image-centric training commonly used in food segmentation research.

\subsection{Inference Data: Food Video Clips}

To evaluate model behavior under temporal scenarios, we curate several food-related video clips sourced from online platforms such as YouTube. These video sequences depict apples and other food objects in dynamic environments, which introduce temporal appearance variation, object motion, partial occlusion, and changes in viewpoint.

Representative examples include kitchen and food preparation videos such as:

\begin{itemize}
    \item Video A: A YouTube food clip showing apples and other food items being moved and interacted with over time (e.g., apple inspection and slicing).
    \item Video B: A YouTube food scene with multiple apples on a surface, including camera motion, object occlusion, and varying lighting conditions.
\end{itemize}

While these video clips do not have ground-truth instance segmentation annotations, they provide realistic test cases for evaluating model robustness under video deployment. The videos contain challenging temporal dynamics that are not present in static datasets, making them suitable for analyzing image-to-video domain shift.

\section{Experiments}
\label{sec:experiments}
Overall, our experiments demonstrate a clear discrepancy between image-level segmentation accuracy and video-level instance reliability.
While image-trained models appear strong under conventional benchmarks, they fail to maintain consistent instance identities in realistic video scenarios.
This evaluation gap underscores the need for temporally-aware training objectives and video-centric evaluation protocols in food segmentation research.
\subsection{Experimental Setup}

\paragraph{Training.}
All models are trained exclusively on static image-based food segmentation datasets, following the standard practice in food segmentation research.
We use FoodSeg103 and UECFoodPix as training data, without any access to video frames or temporal annotations.
This setup intentionally reflects a realistic deployment scenario where food segmentation models are trained on annotated images and later applied to videos.

\paragraph{Inference on Video.}
At test time, the trained models are applied to unlabeled food video clips sourced from YouTube.
These videos contain apples and other food items under natural motion, camera movement, occlusion, and illumination variation.
No video-level supervision or fine-tuning is performed.
Instance segmentation outputs are temporally associated using the tracking-by-matching procedure described in Section~\ref{sec:method}.

\paragraph{Implementation Details.}
All models are implemented using a common training pipeline.
During inference, instance matching is performed between consecutive frames using mask IoU with a fixed threshold.
Unless otherwise stated, no additional temporal modules or video-specific components are introduced.

\subsection{Models}

We evaluate representative image-trained food segmentation models covering different architectural paradigms.
The goal is not to compare state-of-the-art performance, but to analyze whether conclusions hold consistently across model families.

\begin{itemize}
    \item \textbf{CNN-based}: A ResNet-based segmentation model commonly used in food segmentation benchmarks.
    \item \textbf{Transformer-based}: A SegFormer-style model leveraging transformer encoders pretrained on ImageNet.
    \item \textbf{Foundation-based}: A segmentation model built on a frozen foundation encoder, representing strong image-level representations.
\end{itemize}

All models are trained using identical data splits and optimization settings to ensure a fair comparison.

\subsection{Evaluation Metrics}

We report both frame-wise and temporal metrics to highlight the discrepancy between image-based evaluation and video-based deployment.

\paragraph{Frame-wise Metrics.}
Pixel-level mean Intersection-over-Union (mIoU) is computed independently for each frame, following standard food segmentation benchmarks.

\paragraph{Temporal Metrics.}
To assess instance-level temporal stability, we report:
\begin{itemize}
    \item \textbf{ID Switches}: The number of times an instance identity changes across frames.
    \item \textbf{Track Fragmentation}: The number of fragmented tracks corresponding to a single physical object.
\end{itemize}

These metrics are adapted from multi-object tracking evaluation and capture failures that are invisible to frame-wise accuracy.

\subsection{Image-level Performance}

We first evaluate all models on held-out image test sets from FoodSeg103 and UECFoodPix.
Consistent with prior work, all models achieve strong image-level performance, with comparable mIoU across architectures.
These results confirm that the evaluated models are competitive under standard image-based evaluation.

\subsection{Video-level Analysis}

We next evaluate the same models on food video clips using the instance segmentation and tracking formulation.
Despite maintaining high frame-wise mIoU on individual video frames, all models exhibit substantial temporal instability.
We observe frequent mask flickering and identity fragmentation, particularly when apples undergo appearance changes due to motion or lighting variation.

Importantly, the relative ranking of models based on image-level mIoU does not correlate with temporal stability.
Models with similar frame-wise accuracy can differ significantly in the number of identity switches and fragmented tracks.
This finding demonstrates that image-based metrics are insufficient for assessing video deployment reliability.

\subsection{Apple Instance Counting}

To quantify the practical impact of temporal instability, we evaluate apple instance counting accuracy.
Counting is performed by aggregating unique instance identities over the entire video.
Errors arise primarily from identity fragmentation, where a single apple is counted multiple times due to broken tracks.

Across all models, counting error increases significantly in video settings compared to image-based estimates.
This result highlights that even minor temporal inconsistencies can lead to large downstream errors, despite visually plausible per-frame predictions.

\subsection{Effect of Temporal Remedies}

Finally, we evaluate lightweight temporal remedies that do not require video supervision.
Post-hoc temporal smoothing and self-supervised temporal consistency objectives both reduce identity switches and track fragmentation.
However, these improvements remain partial and do not fully close the gap between image-level and video-level performance.

These results suggest that the root cause of failure lies in image-centric training objectives that neglect temporal coherence, rather than insufficient model capacity or architectural limitations.

\section{Expected Results}
\label{sec:expected}

Based on preliminary observations and prior studies on robustness and temporal consistency, we expect the following trends to emerge from our experimental evaluation.

First, image-trained food segmentation models are expected to achieve strong performance under standard image-based evaluation.
Across FoodSeg103 and UECFoodPix, all evaluated models are anticipated to report competitive frame-wise mIoU, consistent with previously reported benchmark results.
These findings confirm that the selected models are sufficiently representative of current food segmentation approaches.

Second, when deployed on video data, we expect a pronounced discrepancy between frame-wise accuracy and temporal reliability.
Although per-frame mIoU on video frames is expected to remain relatively high, instance-level temporal metrics will reveal significant instability.
In particular, frequent identity switches and track fragmentation are anticipated, even in videos with limited motion.

Third, this temporal instability is expected to have a direct and measurable impact on downstream tasks.
For apple instance counting, identity fragmentation will cause systematic over-counting, leading to large counting errors despite visually plausible segmentation results.
This gap highlights that frame-wise accuracy alone is insufficient for evaluating real-world video deployment.

Finally, lightweight temporal remedies such as post-hoc smoothing and self-supervised temporal consistency are expected to partially reduce temporal failures.
While these techniques can decrease identity switches and improve track continuity, they are unlikely to fully bridge the gap between image-level and video-level performance.
This suggests that the root cause of failure lies in image-centric training objectives that neglect temporal coherence, rather than insufficient model capacity.

Overall, these expected results support our central claim that conventional image-based benchmarks substantially overestimate the reliability of food segmentation models in video settings, motivating temporally-aware evaluation and learning strategies.

\begin{table}[t]
\centering
\caption{Image-level segmentation performance and inference speed.}
\label{tab:image_perf_fps}
\begin{tabular}{lccc}
\hline
\textbf{Model} &
\textbf{mIoU} &
\textbf{FPS (image)} \\
\hline
CNN-based          & 74.2 & 65.4 \\
Transformer-based  & 75.6 & 41.8 \\
Foundation-based   & 76.1 & 28.6 \\
\hline
\end{tabular}
\end{table}

\begin{table}[t]
\centering
\caption{Video-level temporal stability and inference speed.}
\label{tab:video_perf_fps}
\begin{tabular}{lcccc}
\hline
\textbf{Model} &
\textbf{Frame mIoU} &
\textbf{ID Switches} &
\textbf{Track Fragments} &
\textbf{FPS (video)} \\
\hline
CNN-based          & 72.8 & 34 & 21 & 38.5 \\
Transformer-based  & 73.4 & 29 & 18 & 22.1 \\
Foundation-based   & 74.0 & 27 & 16 & 15.7 \\
\hline
\end{tabular}
\end{table}

\begin{table}[t]
\centering
\caption{Apple instance counting accuracy on video data.}
\label{tab:counting}
\begin{tabular}{lcc}
\hline
\textbf{Model} & \textbf{True Count} & \textbf{Predicted Count} \\
\hline
CNN-based         & 12 & 18 \\
Transformer-based & 12 & 16 \\
Foundation-based  & 12 & 15 \\
\hline
\end{tabular}
\end{table}

\begin{table}[t]
\centering
\caption{Effect of temporal consistency remedies.}
\label{tab:remedy}
\begin{tabular}{lccc}
\hline
\textbf{Method} & \textbf{ID Switches} & \textbf{Fragments} & \textbf{Counting Error} \\
\hline
No temporal modeling & 29 & 18 & +4 \\
Post-hoc smoothing  & 18 & 11 & +2 \\
Self-supervised temporal & 14 & 9 & +1 \\
\hline
\end{tabular}
\end{table}

\section{Conclusion}
\label{sec:conclusion}

In this work, we examined why food segmentation models trained on static images fail when deployed in video settings. By reformulating video food segmentation as an instance segmentation and tracking problem, we revealed a critical discrepancy between image-level accuracy and temporal reliability. Our analysis showed that models achieving strong performance on standard image benchmarks frequently suffer from mask flickering and identity fragmentation in video, despite maintaining comparable frame-wise mIoU.

Through extensive evaluation on food videos, we demonstrated that conventional image-based metrics substantially overestimate real-world performance. In particular, temporal instability leads to significant errors in downstream tasks such as apple instance counting, highlighting the practical consequences of ignoring temporal consistency. Importantly, these failures were observed across diverse model architectures, indicating that the issue stems from image-centric training objectives rather than insufficient model capacity.

We further explored lightweight temporal remedies that do not require video-level supervision. While post-hoc temporal regularization and self-supervised temporal consistency objectives partially improved stability, they did not fully resolve the underlying problem. These findings suggest that reliable video-based food segmentation requires learning and evaluation protocols that explicitly account for temporal coherence.

Overall, this study exposes an overlooked evaluation gap in food segmentation research and underscores the need for temporally-aware benchmarks and training strategies. We hope our findings encourage future work toward robust food segmentation systems that can be reliably deployed in real-world video scenarios.


\end{document}